\documentclass[letterpaper, 10 pt, conference]{ieeeconf}  

\IEEEoverridecommandlockouts                              
\usepackage{amsmath,amssymb,amsfonts}

\usepackage{subcaption}

\usepackage{amsmath} 
\usepackage{amssymb}  
\usepackage{balance}
\usepackage{graphicx}
\usepackage{times}
\usepackage{dsfont}
\usepackage{multicol}
\usepackage{siunitx}
\usepackage[bookmarks=true]{hyperref}
\usepackage{cleveref}
\usepackage{xspace}
\usepackage{xurl}
\usepackage{capt-of}
\usepackage[usenames,dvipsnames,table,xcdraw]{xcolor}

\usepackage{booktabs}  
\usepackage{float} 
\usepackage[flushleft]{threeparttable}
\usepackage{arydshln} 

\usepackage{caption}
\usepackage{stfloats}

\usepackage[
    style=ieee,
    natbib=true,
    citestyle=numeric-comp,
    doi=false,
    isbn=false,
    url=false]{biblatex} 
\definecolor{mygreen2}{RGB}{0 205 0}

\hypersetup{
	colorlinks=true,
	linkcolor=blue,
	urlcolor=magenta,
	citecolor=mygreen2,
}

\newcommand{\method}{\texttt{HDMI}}
\newcommand{\methodfull}{\textbf{\underline{H}umanoi\underline{D} i\underline{M}itation for \underline{I}nteraction}}

\newcommand{\resterm}{\textbf{w/ residual action, w/ track term}}
\newcommand{\noresterm}{\textbf{w/o residual action, w/ track term}}
\newcommand{\noresnoterm}{\textbf{w/o residual action, w/o track term}}

\newcommand{\rewterm}{\textbf{w/ interaction rew, w/ contact term}}
\newcommand{\norewterm}{\textbf{w/o interaction rew, w/ contact term}}
\newcommand{\norewnoterm}{\textbf{w/o interaction rew, w/o contact term}}

\title{\LARGE \bf
HDMI: Learning Interactive Humanoid \\ Whole-Body Control from Human Videos
}

\author{Haoyang Weng, Yitang Li, Nikhil Sobanbabu, Zihan Wang, \\
Zhengyi Luo, Tairan He, Deva Ramanan, and Guanya Shi\thanks{The authors are with the Robotics Institute, Carnegie Mellon University, USA. \{hweng2, yitangl, nsobanba, zihanwa3, zhengyiluo5, tairanhe, deva, guanyas\}@andrew.cmu.edu}
}

\begin{document}

\IEEEaftertitletext{{
            \vspace{-4mm}
            \centering
            \includegraphics[width=0.99\linewidth]{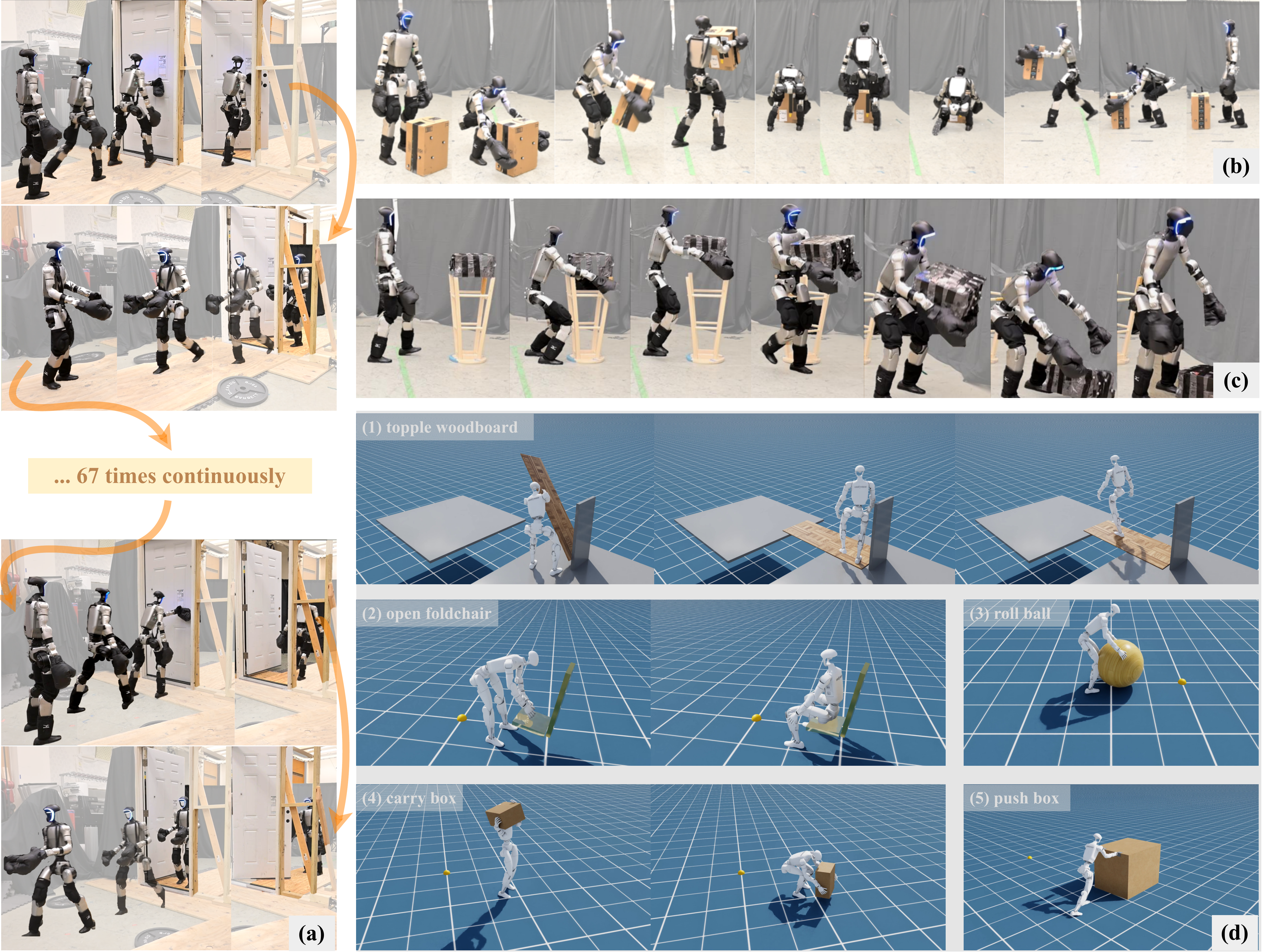}
            \captionof{figure}{\method{} enables humanoid robots to acquire diverse whole-body interaction skills directly from human videos. \textbf{(a)} Traversing doors: the robot successfully passes through a door for \emph{67 consecutive trials} ($\sim 34$ mins), and remains robust under terrain changes. \textbf{(b)} Moving a cardboard box: the robot kneels to grasp and relocate the box, demonstrating coordinated whole-body motion. \textbf{(c)} Carrying and dropping objects: the robot walks forward to pick up and drop a pile of foam mats. \textbf{(d)} A wide range of interaction tasks in simulation, including toppling a wood board, opening a foldable chair, rolling a ball, carrying a box, and pushing a box. 
            Website: \url{https://hdmi-humanoid.github.io}
            }
            \label{fig:teaser}
        }}

\vspace{-6mm}
\maketitle

\begin{abstract}


    Enabling robust whole-body humanoid–object interaction (HOI) remains challenging due to motion data scarcity and the contact-rich nature. We present \method{} (\methodfull{}), a simple and general framework that learns whole-body humanoid-object interaction skills directly from monocular RGB videos. Our pipeline (i) extracts and retargets human and object trajectories from unconstrained videos to build structured motion datasets, (ii) trains a reinforcement learning (RL) policy to co-track robot and object states with three key designs: a unified object representation, a residual action space, and a general interaction reward, and (iii) zero-shot deploys the RL policies on real humanoid robots.
    Extensive sim-to-real experiments on a Unitree G1 humanoid demonstrate the robustness and generality of our approach: \method{} achieves 67 consecutive door traversals and successfully performs 6 distinct loco-manipulation tasks in the real world and 14 tasks in simulation. Our results establish \method{} as a simple and general framework for acquiring interactive humanoid skills from human videos.
\end{abstract}


\clearpage


\section{Introduction}

Humanoid robots hold immense potential for assisting humans in diverse environments due to their human-like morphology and versatility. To fully unleash their capabilities, enabling humanoids to robustly interact with objects and their environments is critical.

Learning from human motions has been a dominant pipeline in humanoid robotics, which leverages abundant human data to achieve agile locomotion \cite{he2025asapaligningsimulationrealworld,beyond_mimic,ji2024exbody2} and dexterous manipulation \cite{zhao2024-aloha-unleashed, ghosh2024-octo, wang2024-HPT}. 
However, there are significant limitations when extending these successes to whole-body, contact-rich humanoid-object interaction (HOI) tasks. We identify two key challenges: (1) Compared to free-space locomotion, human-object interaction data with 3D human and object motions is scarce. (2) Learning whole-body interaction tasks poses new challenges for RL training, such as guiding desired contact behavior under imperfect motion references and learning to balance with objects in challenging poses.


Prior humanoid-object interaction works have either relied on task-specific motion reference generation pipelines or manual reward engineering~\cite{opt2skill, dao2023simtoreallearninghumanoidbox} that limit generality or require high-level VLM or model-based planners~\cite{qiu2024-wildlma, su2025hitterhumanoidtabletennis}. To overcome these limitations, we propose \method{} (\methodfull{}), a general framework for acquiring interactive skills directly from RGB videos. Our framework supports interaction with different body parts (hand, foot), and with different types of objects (articulated or rigid body, fixed-based, floating-based).
Our key insight is to bypass task-specific reward engineering by jointly tracking robot and object motions from video with an end-to-end RL control policy.
Our pipeline has three stages: (i) extract and retarget human and object trajectories from RGB videos using pose estimation~\cite{shen2024gvhmr, alhafez2023b} to build structured reference datasets; (ii) train a control policy with RL to co-track robot and object states, using proprioception, a phase variable, and object state in the robot’s local frame;
and (iii) deploy the learned policy directly on humanoid robot to execute interaction skills.

Our contribution is three-fold:
\begin{itemize}
\item We propose \method{}, a simple and general framework that learns interactive skills for humanoids directly from RGB videos. As far as we know, this is the first general framework that enables learning autonomous whole-body humanoid-object interaction skills directly from human videos.
\item To facilitate stable and efficient training for complex humanoid-object interactions, we design \textit{three simple and unified components}: a \textit{unified \underline{object representation}} for diverse objects, a \textit{\underline{residual action space}} for stable exploration of challenging poses, and a \textit{unified \underline{interaction reward}} that promotes robust and precise contact even with imperfect reference motions.
\item We demonstrate the robustness and generality of our framework through extensive sim-to-real experiments on Unitree G1. As shown in \Cref{fig:teaser}, the learned policy achieves \textit{\underline{67 consecutive} bi-directional door openings and traversals}, and we successfully train and deploy other \textit{\underline{6 distinct loco-manipulation tasks}} on real hardware and \textit{\underline{14 tasks}} in simulation. 
\end{itemize}


\section{Related Works}

\subsection{Humanoid Learning}
RL has made remarkable progress for agile humanoid locomotion skills. Through large-scale training, humanoids have achieved robust walking and running
\cite{ li2024reinforcement, gu2024advancing}
, dynamic behaviors such as jumping~\cite{li2023robust}, parkour~\cite{zhuang2024humanoid}, and expressive whole-body motions~\cite{ji2024exbody2, cheng2024expressive}. In parallel, imitation learning and large-scale human demonstrations have produced dexterous manipulation policies~\cite{zhao2024-aloha-unleashed, ghosh2024-octo, wang2024-HPT}, enabled by advances in data collection~\cite{arunachalam2023dexterous, chi2024universal, humanpoli-humanoidpoli}. Despite this progress in locomotion and manipulation individually, relatively few works have addressed humanoid loco-manipulation, where the robot must simultaneously move and interact with objects in contact-rich settings~\cite{homie,he2024learning, he2024omnih2o}. Our work focuses on this challenging frontier, developing policies that couple locomotion with object interaction to enable robust, and expressive humanoid skills in the real world.

\subsection{Humanoid Loco-Manipulation}
Learning-based methods have recently extended from locomotion and manipulation in isolation to full humanoid loco-manipulation, yet existing approaches remain limited in generality and robustness. Some rely on reward engineering or trajectory optimization, achieving task-specific successes such as box transport~\cite{opt2skill, dao2023simtoreallearninghumanoidbox}. Others introduce specialized architectures, including skill blending~\cite{kuang2025skillblender} or decouple low-level tracking with high level task policies~\cite{he2024learning, he2024omnih2o, ji2024exbody2}. 
More recent efforts address robustness through dual-agent RL frameworks ~\cite{falcon,jaeger}, and adaptive policies for human–humanoid collaboration~\cite{h2compact}. 
While these advances expand humanoid capabilities, their applicability is relatively narrow.
Our framework instead introduces a general, dense, demonstration-driven objective that couples locomotion with object interaction, enabling adaptive and contact-stable behaviours that scale across diverse tasks.

\subsection{Robot Learning from Human Videos} 
Recent advances in humanoid robot learning have increasingly turned to human videos as a scalable source of demonstration data. Despite the success of methods learning locomotion skills from video demonstrations \cite{he2025asapaligningsimulationrealworld, mao2024learningmassivehumanvideos, wang2025hilhybridimitationlearning, videomimic}, these approaches fundamentally lack object interaction capabilities as they do not explicitly model object dynamics during training. Another line of work focuses on learning manipulation skills from video demonstrations \cite{okami2024, zhu2024vision, zhou2025teachoncelearnoneshot}. However, these methods are typically constrained to only upper-body interactions, lacking potential to utilize the large workspace that can be achieved with whole body coordination. To address this limitation, \method{} learns whole-body interactions from monocular RGB videos, modeling human–object dynamics and co-tracking trajectories to unify locomotion and manipulation.

\clearpage

\begin{figure*}[t] 
    \centering
    \includegraphics[width=0.95\linewidth]{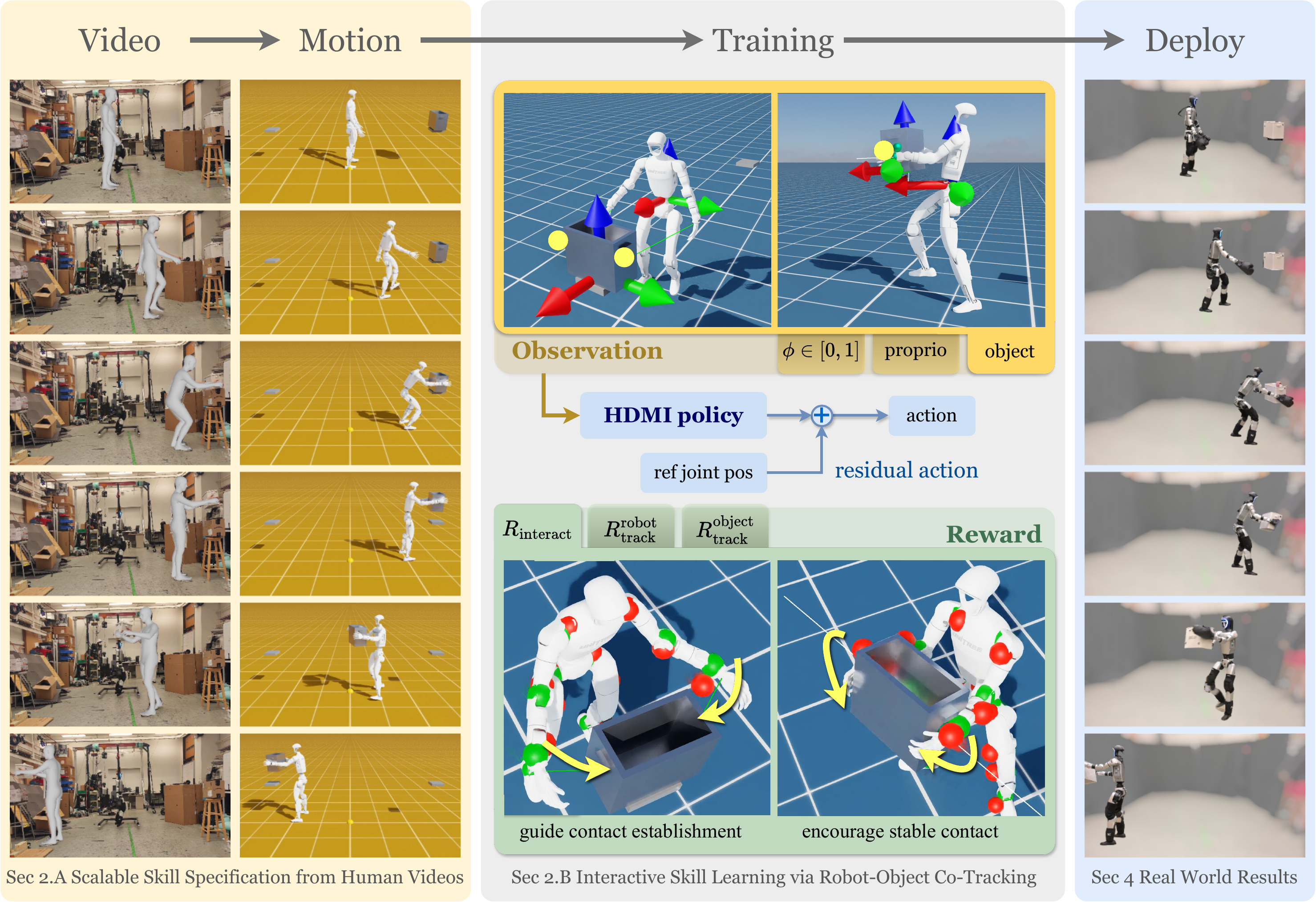}
    \captionof{figure}{\method{} is a general framework for interactive skill learning. 
    Monocular RGB videos are processed into a structured dataset as reference trajectories (\cref{sec:method-retarget}), which are used to train an interaction centric policy via robot-object co-tracking (\cref{sec:method-tracking}). The trained policies are succesfully deployed to real world humanoids (\cref{sec:real_world}).}
    \label{fig:method_pipeline}
\end{figure*}
\vspace{-0.1in}

\section{Method}
\method{} is a general framework for interactive skill acquisition. 
We first retarget and process monocular RGB videos of human interacting with objects into a structured dataset of reference trajectories (\cref{sec:method-retarget}). Then these reference trajectories are used to train a interaction centric policy via robot-object co-tracking (\cref{sec:method-tracking}).


\subsection{Scalable Skill Specification from Human Videos}
\label{sec:method-retarget}
A primary challenge in teaching robots complex interaction skills is the difficulty of specifying the desired behavior. 
To address this, we leverage monocular RGB videos of humans interacting with objects as an abundant and scalable data source. We use GVHMR \cite{shen2024gvhmr} and LocoMujoco \cite{alhafez2023b} for smpl-pose estimation and retargeting. Next, we post-process and annotate the object trajectory and contact signals to produce a structured dataset of reference motions, in which each frame provides a reference state $\{\mathbf{s}_t^{\text{ref}}\}$ together with desired contact points defined in the object’s local frame $\{\mathbf{p}^{\text{contact}}_t\}$.
The reference state at frame $t$ is $\mathbf{s}_t^{\text{ref}} = (\mathbf{s}_t^{\text{robot}}, \mathbf{s}_t^{\text{obj}}, \mathbf{c}_t)$ where 
$\mathbf{s}^{\text{robot}}$ consists of $\mathbf{p}^{\text{robot}}$, $\mathbf{q}^{\text{robot}}$ and $\mathbf{\theta}^{\text{ref}}_t$, representing the robot’s reference root position, root orientation and reference joint positions. 
$\mathbf{s}^{\text{obj}}$ consists of $\mathbf{p}^{\text{obj}}$ and $\mathbf{q}^{\text{obj}}$, representing the object’s position and orientation. For articulated objects (e.g., doors or folding chairs), $\mathbf{s}_t^{\text{obj}}$ additionally includes their joint state $\mathbf{\theta}^{\text{obj}}$.
$\mathbf{c}_t \in \{0, 1\}$ is a binary signal indicating whether contact is intended at frame $t$.


\subsection{Interactive Skill Learning via Robot-Object Co-Tracking}
\label{sec:method-tracking}

We formulate interactive skill learning as a robot–object co-tracking problem. Specifically, given a processed reference motion $\{\mathbf{s}^{\text{ref}}_t, \mathbf{p}^{\text{contact}}_t\}$, our goal is to train a whole-body control policy that simultaneously follows the reference trajectory of both the robot and the object at each timestep using reinforcement learning.


Following established tracking works \cite{peng2018deepmimic, he2025asapaligningsimulationrealworld, beyond_mimic}, we use a DeepMimic-style~\cite{peng2018deepmimic} training: 
(1) \textit{Reference state initialization}: during each episode, both the robot and object are initialized from a random frame $s_t^{\text{ref}}$ in the reference motion, with additional small random perturbations to enhance robustness.
(2) \textit{Phase variable observation}: policy receives a phase variable $\phi \in [0, 1]$, where $\phi = 0$ represents the start of a motion and $\phi = 1$ represents the end. This time phase variable alone is proven to be sufficient for single-motion tracking~\cite{peng2018deepmimic,he2025asapaligningsimulationrealworld}.
(3) \textit{Tracking-error-based termination}: we terminate the episode when robot or object states deviate too much from the reference motion. We train the policy with tracking and regularization rewards and use PPO to optimize it. A detailed list of rewards and terminations can be found in \cref{tab:rewards_unified} and \cref{tab:termination_conditions}.

However, learning robust whole-body object interaction propose new challenges. We address the key challenges by introducing three targeted solutions:

\subsubsection{\textbf{Unified Object Representation}}
\label{sec:method-tracking-obs}

To allow our framework to generalize across diverse objects with different geometries and types,
we design a unified representation for object observations.
At each timestep, the policy receives the object’s pose expressed relative to the robot’s local root frame. This spatially invariant formulation facilitates generalization and can be naturally distilled to on-board sensory inputs such as RGB or depth images.

\begin{figure}[h!]
    \centering
    \begin{subfigure}[b]{0.32\linewidth}
        \centering
        \includegraphics[width=\linewidth]{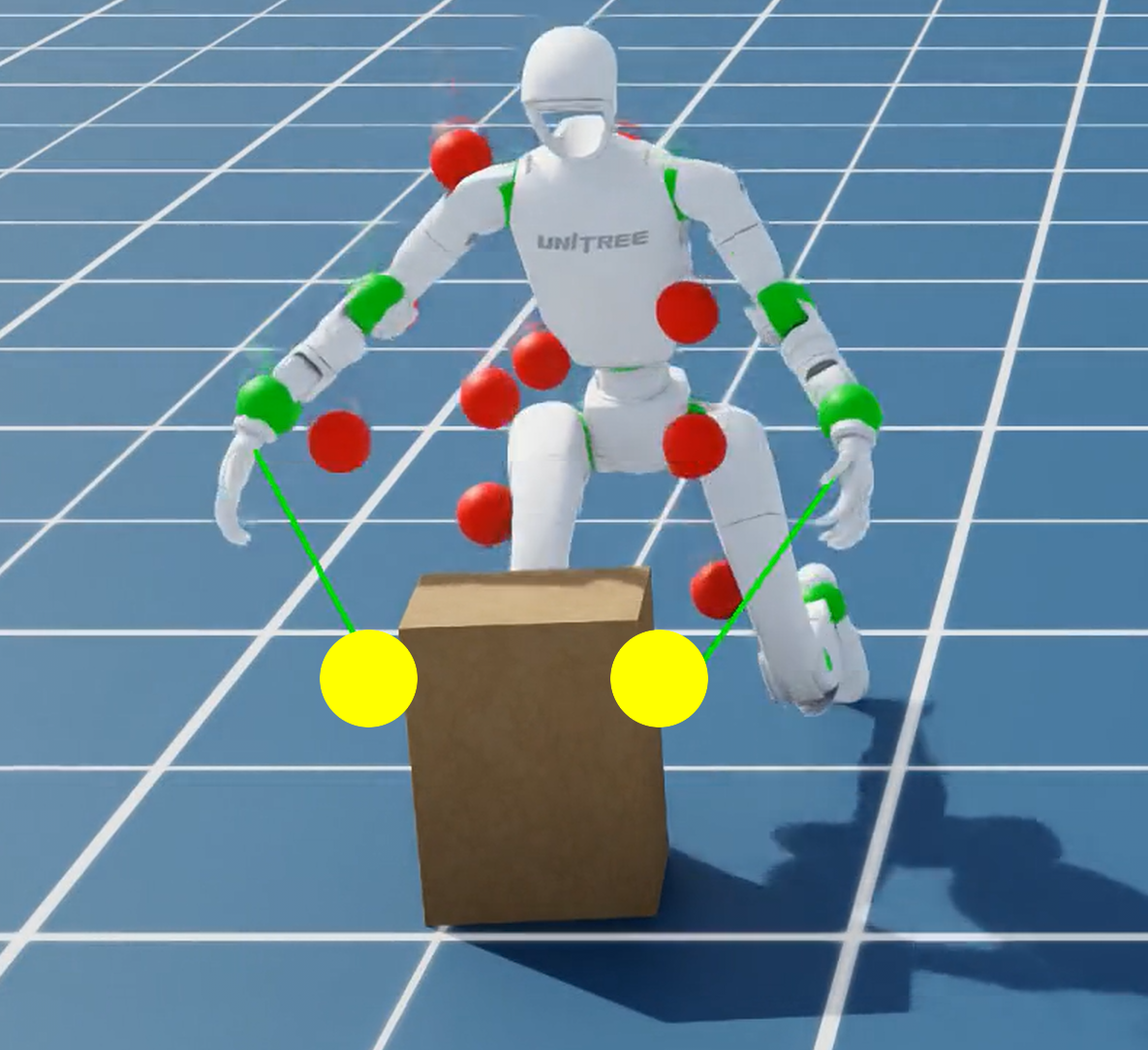}
    \end{subfigure}
    \begin{subfigure}[b]{0.32\linewidth}
        \centering
        \includegraphics[width=\linewidth]{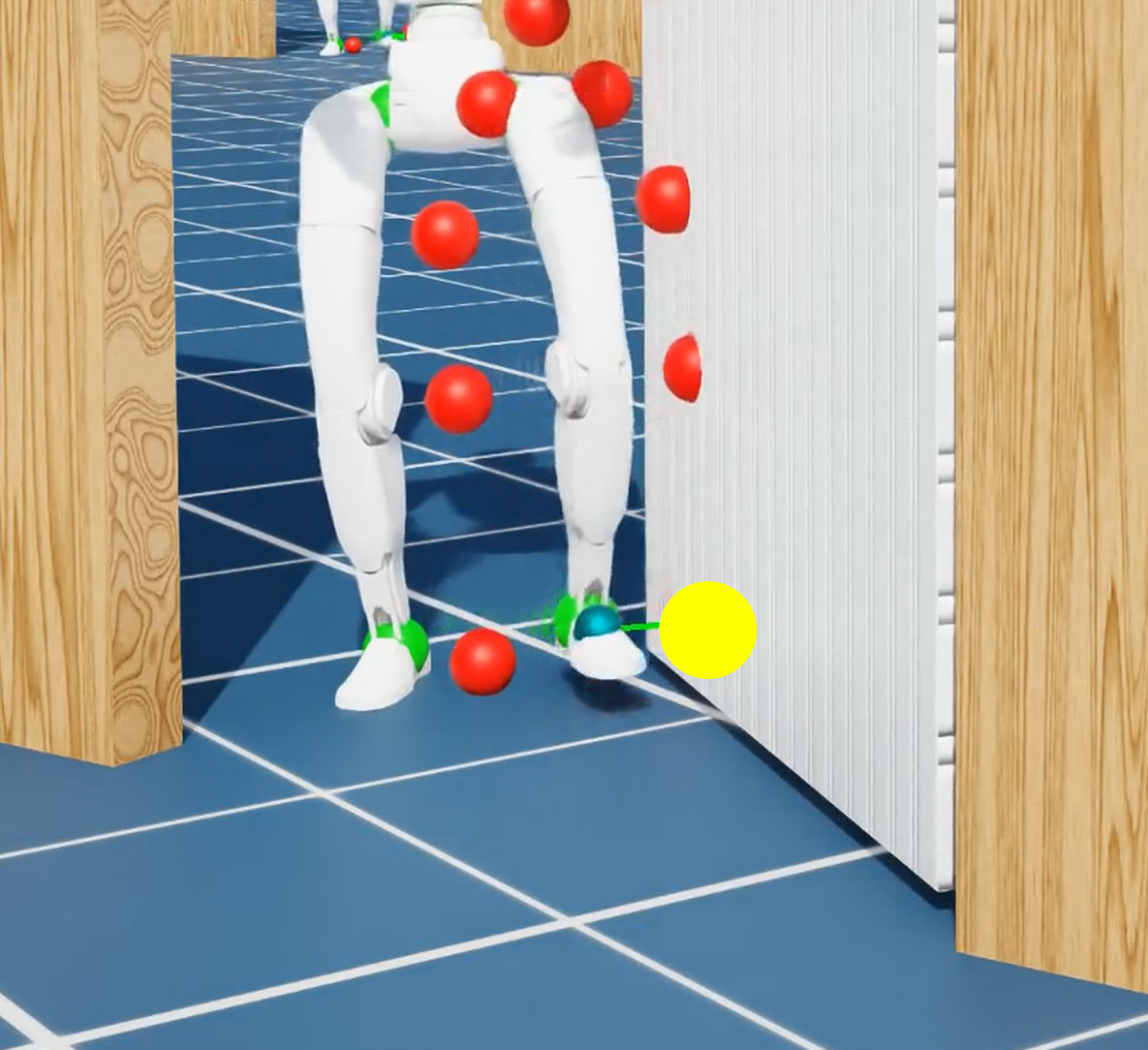}
    \end{subfigure}
    \begin{subfigure}[b]{0.32\linewidth}
        \centering
        \includegraphics[width=\linewidth]{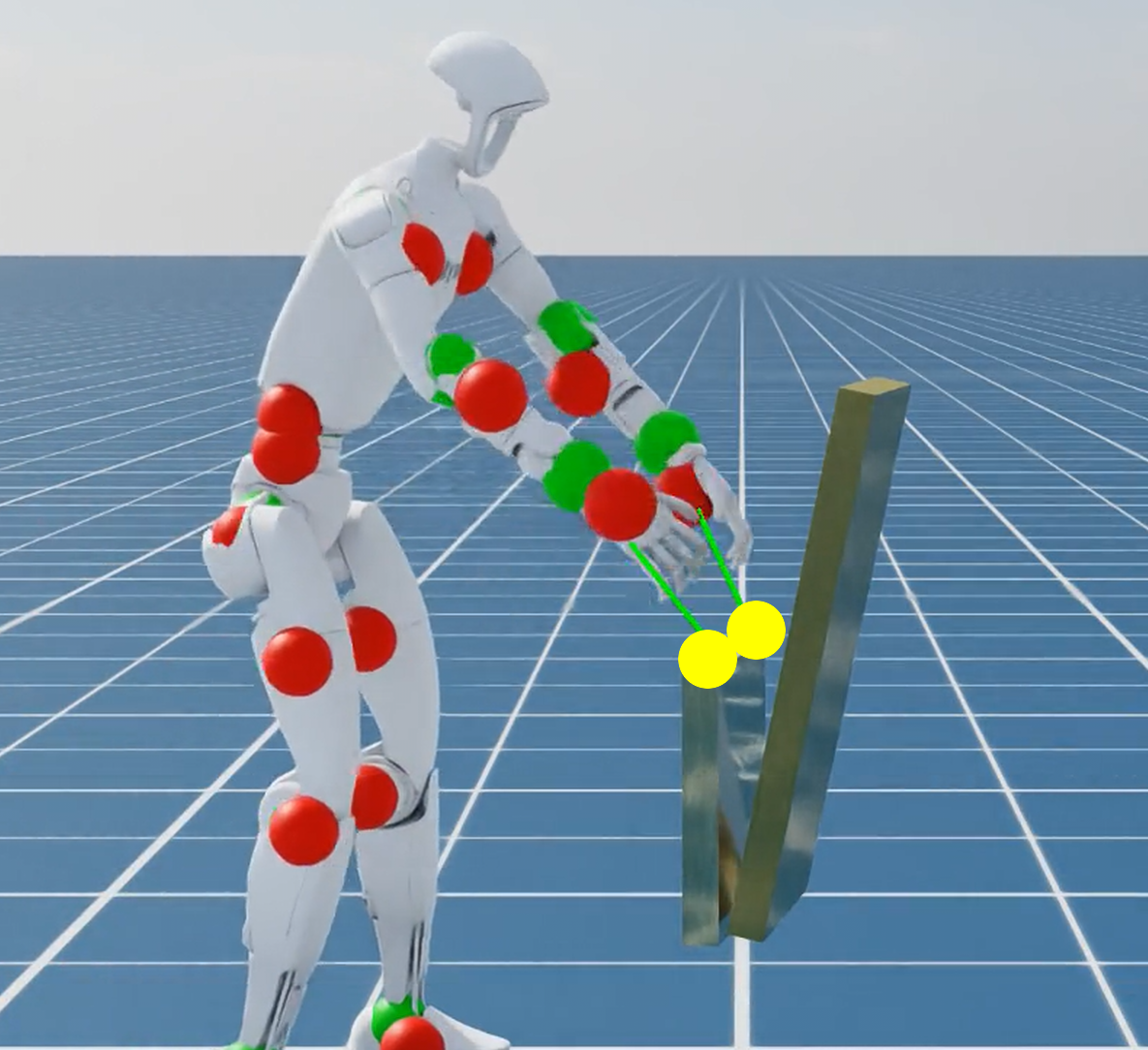}
    \end{subfigure}
    \caption{Reference contact position (yellow dot) in three different tasks. Policy observes the positions of these contact points in root frame, both during training and deployment.}
    \label{fig:method-uni_obj_obs}
\end{figure}

To further guide the interaction, we also provide the policy with reference contact points $\mathbf{p}^{\text{contact}}$, which specify desired robot–object contact locations as shown in \cref{fig:method-uni_obj_obs}. These points are also transformed into the robot’s root frame. Together with the robot’s proprioceptive state $\mathbf{s}_t^{\text{proprio}}$ and the phase variable $\phi \in [0,1]$, this unified observation (\cref{fig:method_pipeline} top) allows our framework to be applied to different object types without architectural changes.

\subsubsection{\textbf{Residual Action Space}}
\label{sec:method-tracking-action}
Learning complex motions often involves challenging poses, such as kneeling, that are far from the robot's default standing pose. At the beginning of training, the exploration of a standard policy is centered around the default pose. When an episode is initialized in a kneeling reference pose, the policy's first action often causes the robot to abruptly pop up toward a standing pose, leading to immediate instability and uninformative training samples.

To overcome this, we employ a residual action space. Instead of learning the absolute joint target $\theta_t^{\text{target}}$ directly, the policy learns to output a corrective offset $\mathbf{a}_t$, which is added to the joint positions from the reference kinematic trajectory $\theta_t^{\text{ref}}$ (\cref{fig:method_pipeline} middle) forming: 
$\theta_t^{\text{target}} = \theta_t^{\text{ref}} + \mathbf{a}_t$.
This grounds the initial exploration to be centered around the current reference pose. For example, if the reference motion kneels down on the ground, due to exploration in the residual action space, the policy learns how to balance itself, while in the direct action space, the policy needs to learn to output large offsets to the zero pose.
This targeted exploration leads to improved sample efficiency and significantly faster convergence to the desired behavior, especially for complex motions that deviate significantly from a standard standing pose.

\subsubsection{\textbf{Unified Interaction Reward}}
\label{sec:method-tracking-reward}
Reference trajectories obtained via video retargeting are purely kinematic and often lack precise contacts or contain penetration artifacts. Relying solely on motion-tracking rewards is therefore insufficient. To address this gap, we introduce a unified contact-promoting reward $R_{\text{interaction}}$, which encourages the policy to establish and maintain stable contact when the reference indicates the intended interaction ($\mathbf{c}_t = 1$).

For each active end-effector $i$, the reward combines (i) a position term that encourages alignment with the target contact point, and (ii) a force term that promotes stable yet bounded contact forces:
\begin{equation}
\begin{split}
    R_{\text{contact}, i} = & \exp\left(-\frac{\|\mathbf{p}_{\text{eef}, i} - \mathbf{p}_{\text{target}, i}\|_2}{\sigma_{\text{pos}}}\right) \cdot \\ 
    &\min\left(\exp\left(\frac{ \|\mathbf{F}_{\text{contact}, i}\|_2 - F_{\text{thres}}}{\sigma_{\text{frc}}}\right), 1\right)
\end{split}
\end{equation}
where $\mathbf{p}_{\text{eef}, i}$ and $\mathbf{p}_{\text{target}, i}$ are the positions of the end-effector and its target contact point on the object, respectively. $\mathbf{F}_{\text{contact}, i}$ is the contact force between the eef and the object.
The position term rewards proximity between end-effector and object contact points.
The force term rewards sufficient but not excessive contact force, capped by a threshold $F_{\text{thres}}$ to ensure safety at deployment.
The overall interaction reward is averaged across all active end-effectors, gated by the contact signal.
\begin{equation}
R_{\text{interaction}} = \left( \frac{1}{N_c} \sum_{i=1}^{N_c} R_{\text{contact}, i} \cdot \mathbf{c}_{t, i} \right)
\end{equation}
where $N_c$ represents the number of end-effectors desired to have contact.


\textit{Domain Randomization:} To improve the robustness of the trained policy, we randomize the robot and object's inertial and friction properties during training.
\vspace{-3mm}
\begin{table}[h!]
\caption{Reward Functions.}
\label{tab:rewards_unified}
\centering
\begin{tabular}{@{}c@{\hspace{0.0em}}c@{}}
\toprule
\begin{tabular}[t]{lc}
\textbf{Reward} & \textbf{Weight} \\
\toprule
\multicolumn{2}{l}{\textbf{Robot Tracking}} \\
\midrule
Body Local Pose             & 2.0 \\
Root Global Pose            & 1.0 \\
Body Global Velocity        & 1.0 \\
Joint Tracking              & 1.0 \\
\midrule
\multicolumn{2}{l}{\textbf{Object Tracking \& Interaction}} \\
\midrule
Object Pose                 & 2.0 \\
Contact Reward              & 5.0 \\
\bottomrule
\end{tabular}
&
\begin{tabular}[t]{lc}
\textbf{Reward} & \textbf{Weight} \\
\toprule
\multicolumn{2}{l}{\textbf{Regularization Penalties}} \\
\midrule
Action Rate L2              & 0.1 \\
Joint Position Limits L1    & 10.0 \\
Joint Velocity Penalty      & 5.0e-4 \\
Joint Torque Limits L1      & 0.01 \\
Feet Impact Force L2        & 1.0 \\
Feet Slip                   & 0.5 \\
Feet Air Time               & 5.0 \\
\bottomrule
\end{tabular}
\end{tabular}
\end{table}
\vspace{-3mm}
\begin{table}[h!]
\caption{Termination Conditions.}
\label{tab:termination_conditions}
\centering
\begin{tabular}{lcc}
\toprule
\textbf{Termination Condition} & \textbf{Threshold} & \textbf{Min Steps} \\
\midrule
\multicolumn{3}{l}{\textbf{Robot Tracking Error}} \\
\midrule
Root Global Pose Error & 0.5m, 1.2rad & 25 \\
Body Local Pose Error & 0.5m, 1.2rad & 25 \\
\midrule
\multicolumn{3}{l}{\textbf{Object Tracking Error}} \\
\midrule
Object Pose Error  & 0.5m, 1.2rad & 25 \\
\midrule
\multicolumn{3}{l}{\textbf{Contact Loss}} \\
\midrule
Lost Contact & Pos 0.2m \& Force 1.0N & 25 \\
\bottomrule
\end{tabular}
\end{table}

\clearpage

\begin{figure*}[t]
    \centering
    \includegraphics[width=0.95 \linewidth]{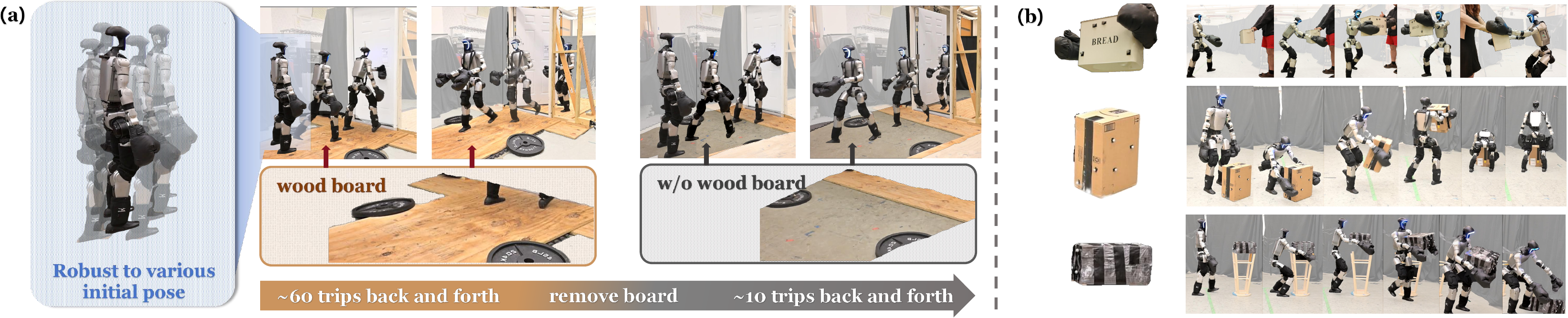}
    \caption{
        Demonstrations on challenging real-world tasks. \textbf{(a)} Door opening and traversal: the robot adapts its footsteps to different initial poses and terrain variations (with/without wooden board), successfully completing 67 consecutive trips. \textbf{(b)} Box loco-manipulation: the policy enables versatile whole-body coordination for grasping, lifting, and transporting objects of varied shapes, sizes, and weights.
    }
    \label{fig:real-door_box}
\end{figure*}
\vspace{-10mm} 


\begin{figure*}[b]
    \centering
    \includegraphics[width=0.95 \linewidth]{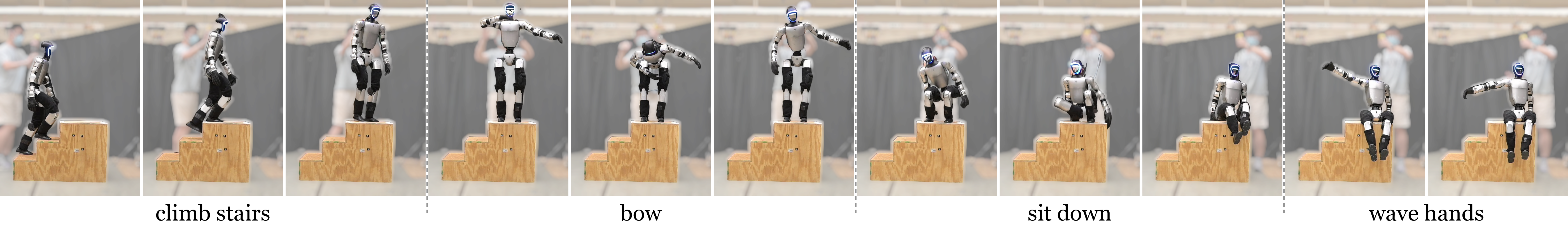}
    \captionof{figure}{\textbf{Truman's Bow.}
    This demonstration highlights long and continuous sequence of diverse, contact-rich behaviors.}
    \label{fig:real-truman}
\end{figure*}

\section{Real-World Experiments}
\label{sec:real_world}


We evaluate \method{} on five real-world interaction tasks.
These tasks collectively test the robot’s ability to handle contact-rich interactions, combine locomotion and manipulation, and perform long-horizon whole-body coordination.

\textbf{Experiment Setup.} We train all policies in IsaacSim with same set of hyperparameters and directly deploy them on a Unitree G1 humanoid. To obtain inputs for the policy, we attach mocap markers on the robot’s pelvis and on the manipulated objects, to acquire the global pose of each root link. Object poses are then transformed into the robot’s root frame to serve as policy observations.

All reference motions are derived from RGB videos, except for suitcase manipulation, which is retargeted from the Omomo dataset \cite{li2023object}. While Omomo offers high-quality human–object interaction data, many motions involve prehensile manipulation (e.g., moving a monitor or lamp) that requires dexterous hands, which is outside the scope of our study.

\subsection{Case I: Door Open and Traversal}

This task tests the policy's ability to handle different contact interactions in sequence. The robot must push a door open with its hand, walk through, turn around, kick the door open with its foot, pass through again, and return to start.

The policy completed 67 continuous door runs (34 minutes) before failure. It remained robust under terrain changes, successfully performing $\sim7$ runs after the wooden floorboard was removed. 
As shown in \cref{fig:real-door_box}(a), the robot started each round with a random positional offset of 10–30 cm. Despite this variation, it adapted its footsteps and precisely raised its arm to make contact at the correct location.

These results highlight the policy's robustness to environmental variation and adaptability to positional changes.

\subsection{Case II: Box Loco-Manipulation}
These tasks evaluate whole-body coordination to manipulate box-like objects of varying heights and weights (\cref{fig:real-door_box} b). The robot must integrate whole-body motions, such as kneeling, grasping, carrying, turning, and placing objects.

\begin{enumerate}
    \item Suitcase manipulation: The robot executed 7 consecutive successful runs with smooth transitions between kneeling, lifting, and walking with load. 
    \item Bread box carrying: The policy completed 2 full trials. The main challenge was the rapid 180° turn, which occasionally caused leg collisions.
    \item Foam mats relocation: The robot walks forward, grasps the mats, sidesteps, and places them down successfully.
\end{enumerate}

These results shows \method{} enables seamless whole-body coordination between grasping, locomotion, and manipulation across various object properties and heights.

\subsection{Case III: Truman's Bow}

This complex, multi-stage task requires executing a long sequence of behaviors: climbing a staircase, performing a Truman-style bow, sitting on the stair, waving, jumping off, and walking back.

Due to the risks of stair climbing, human operators occasionally applied light support to the robot’s back. The learned policy successfully executed the entire sequence 3 times continuously. 
This demonstrates the framework's robust capability for versatile and complex motions, encompassing dynamic and contact rich object-scene interactions (e.g., climbing and sitting on stairs) and precise full-body pose control for expressive gestures (e.g. bowing, waving hands).

This highlights \method{}’s ability to produce highly contact-rich, long-horizon whole-body behaviors in the real world.


\clearpage

\section{Simulation Ablations}
\label{sec:sim_ablation}

We conduct a set of experiments to analyze the contribution of key components and design choices within our framework.
All simulation experiments are conducted in IsaacSim.
Each evaluation is initialized at the start of the reference and is considered successful if policy finishes the task without triggering any termination. Mean and standard deviation of metrics are computed over 4096 parallel simulation environments.

%

\subsection{Interaction Reward}

We study the effectiveness of two design choices that guide learning towards desired contact behavior: \textbf{interaction reward} and \textbf{contact based termination} (\cref{tab:termination_conditions}, \emph{Lost Contact}).
We compare three training variants:
\begin{itemize}
\item \rewterm{}: Incorporates the interaction reward and contact-based termination.
\item \norewterm{}: Only removes the interaction reward.
\item \norewnoterm{}: Removes both.
\end{itemize}
Unless otherwise specified, contact-based terminations are removed for evaluations in this experiment.

\begin{figure}[b]
    \centering
    \includegraphics[width=\linewidth]{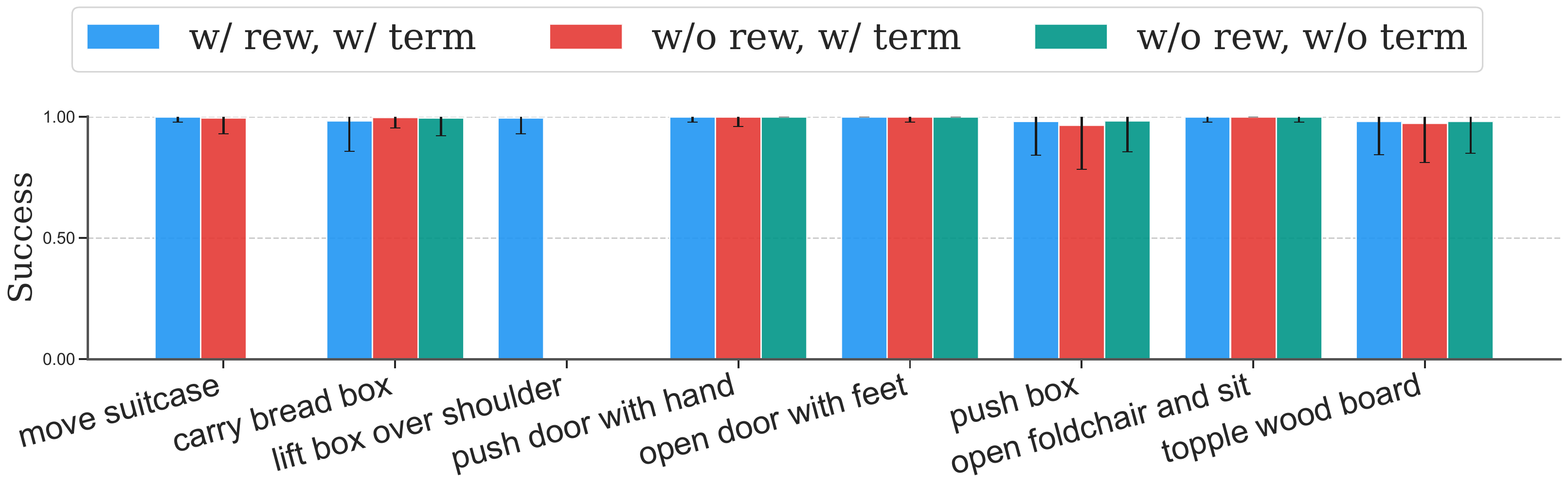}
    \caption{
    \textbf{Final success rate across 8 tasks.} 
    For most tasks removing contact reward and contact based termination actually does not affect final success rate.
    }
    \label{fig:ablation-contact_rew-table}
\end{figure}

For the majority of tasks, removing these two components does not significantly impact the final performance. However, we identified two specific types of tasks where these components are crucial, and provide a detailed analysis below.

\begin{figure}[h!]
    \centering

    \begin{subfigure}{\linewidth}
        \centering
        \includegraphics[width=\linewidth]{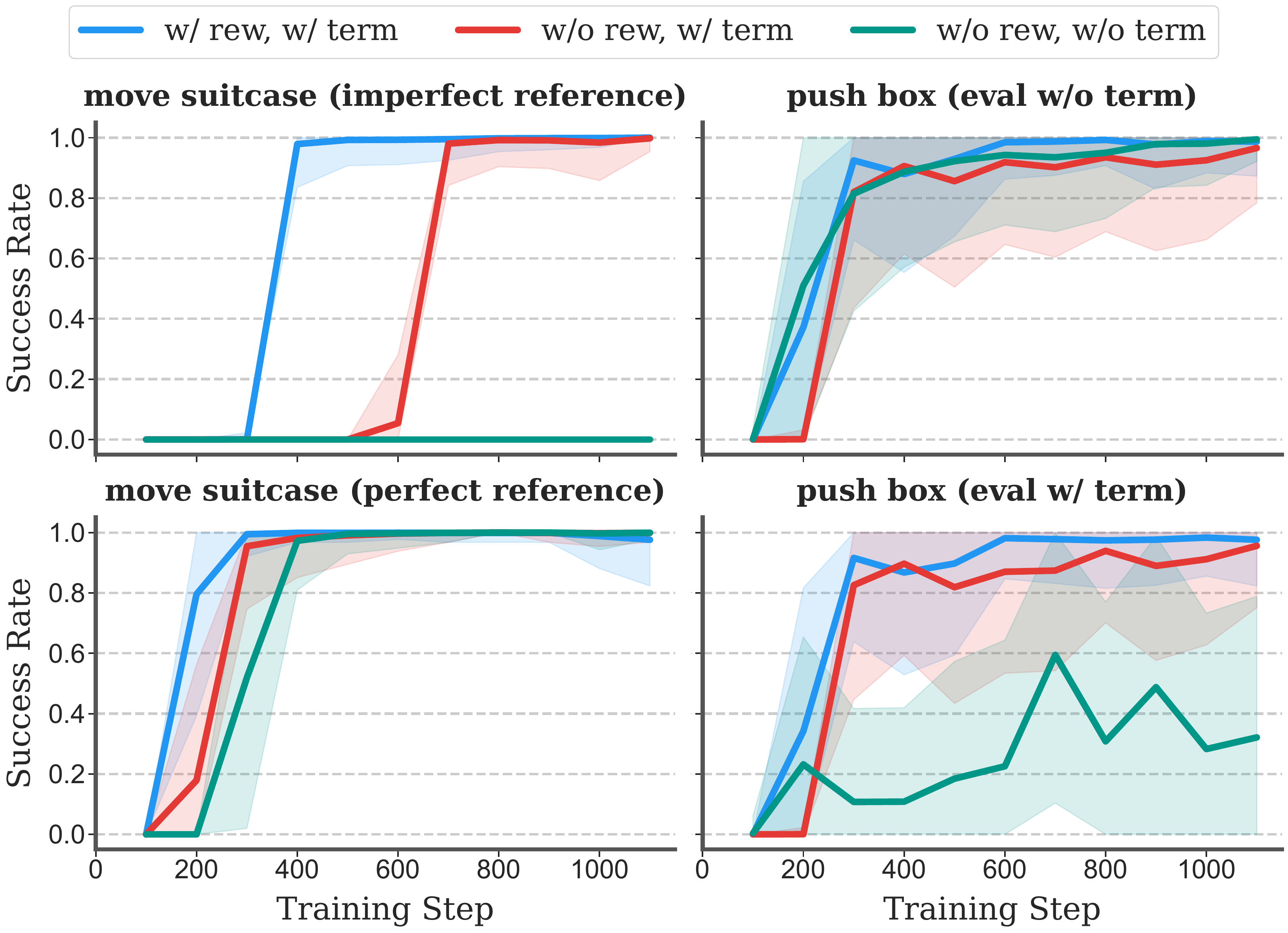}
        \caption{
        (left) For move suitcase, interaction reward is crucial for task success when reference is imperfect. 
        (right) For push box, 
        while all policies show comparable success rates without contact-based termination (eval w/o term), enabling it causes a significant drop for the policy trained without our interaction reward. This reveals that our reward is essential for learning to maintain a stable contact.
        }
        \label{fig:ablation-contact_rew-training}
    \end{subfigure}
    
    \begin{subfigure}{\linewidth}
    \end{subfigure}
    
    \begin{subfigure}{\linewidth}
        \centering
        \includegraphics[width=0.92 \linewidth]{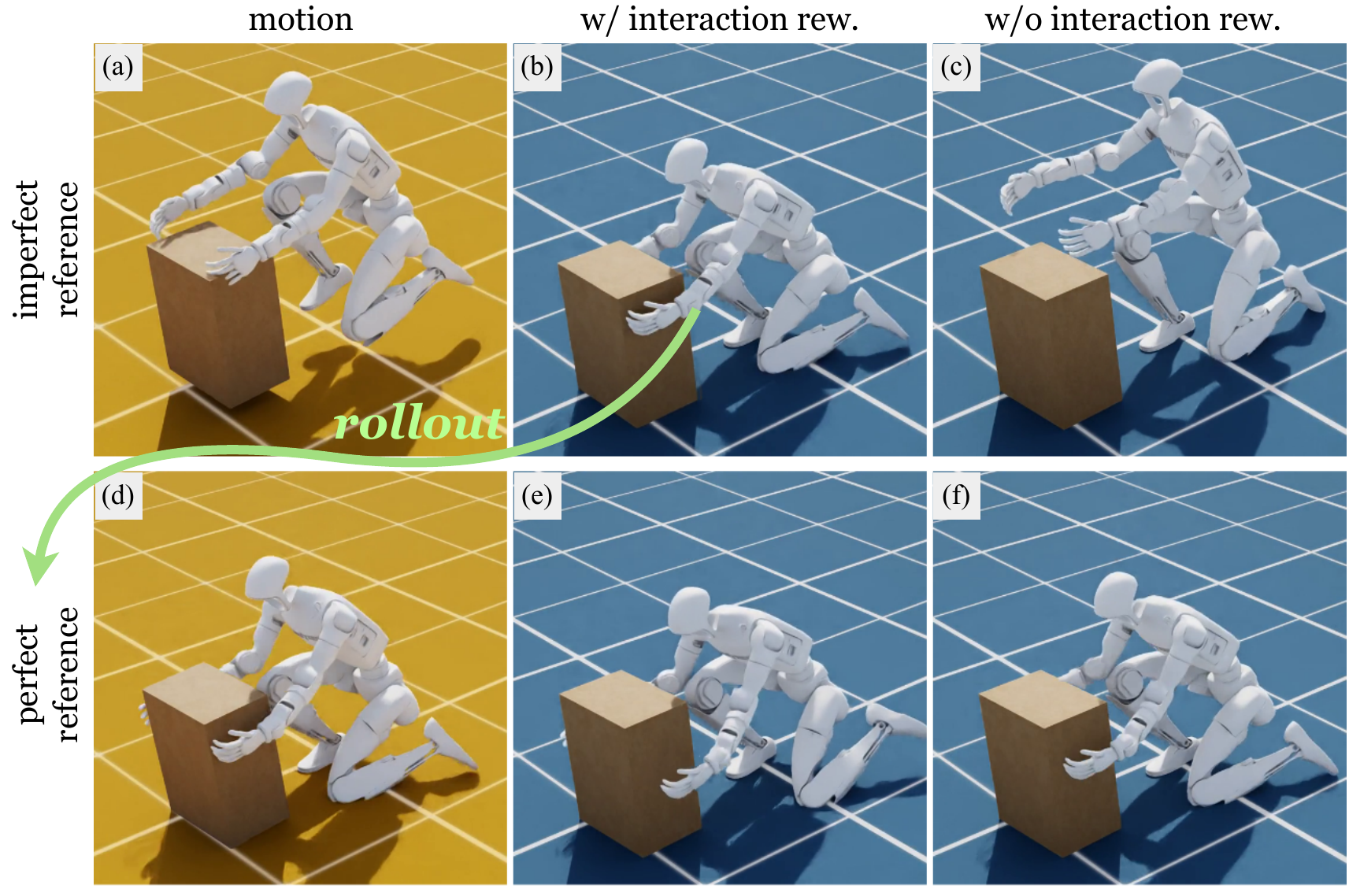}
        \caption{\textbf{Interaction reward guides contact establishment with imperfect reference.} 
        With imperfect motion (a), interaction reward drives the policy to deviate from the flawed reference to establish a grasp (b). Without it, the policy rigidly mimics the reference and fails (c).
        Rollouting (b) we collect a perfect reference (d), interaction reward is not crucial for success when trained with this perfect reference.
        }
        \label{fig:ablation-contact_rew-policy-suitcase}
    \end{subfigure}

    \begin{subfigure}{\linewidth}
        \centering
        \includegraphics[width=0.9 \linewidth]{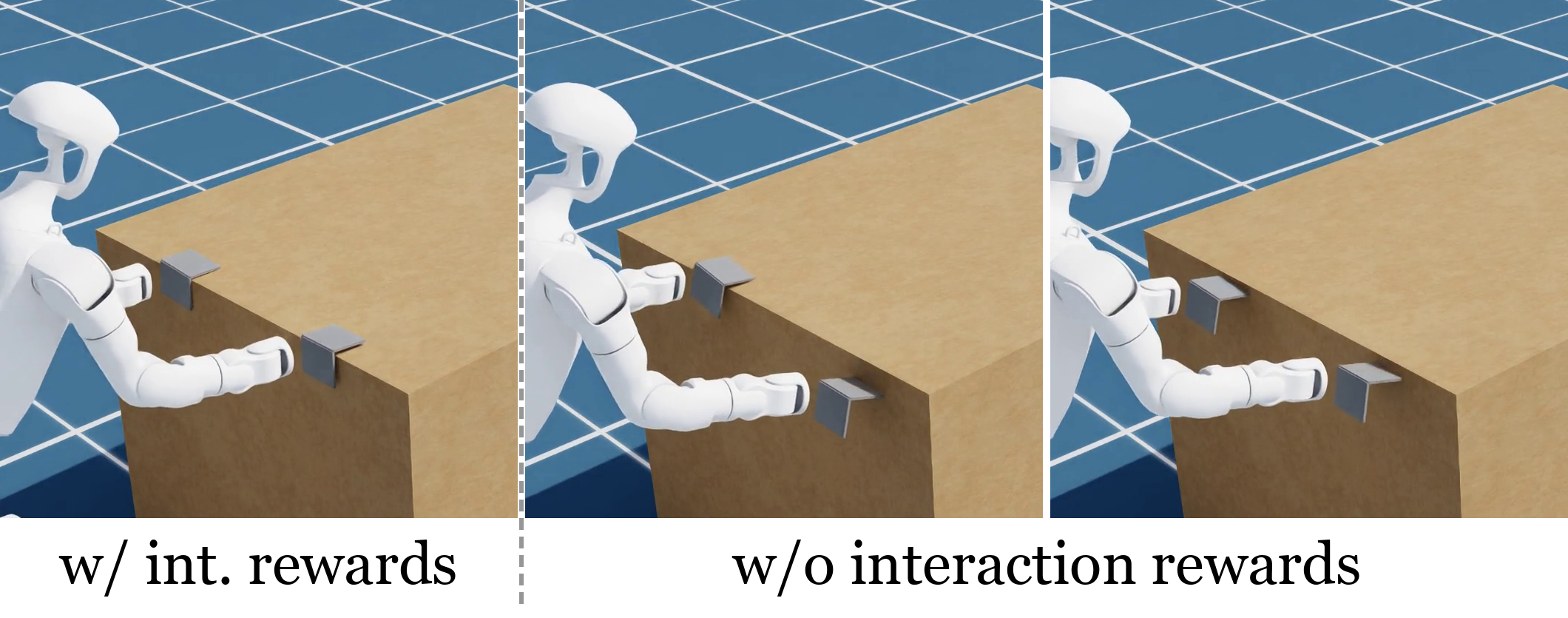}
        \caption{\textbf{Interaction reward guides precise contact locations.} 
        (left) Interaction reward guides the policy to accurately place its L-shaped end effector on the box's edge.
        (right) Without it, the policy fail to achieve precise contact, frequently placing the end effectors on the vertical surface of the box, leading to an unstable interaction.
        }
        \label{fig:ablation-contact_rew-policy-push_box}
    \end{subfigure}
    \caption{Ablation study on interaction reward and contact-based termination.}
\end{figure}

\textbf{Interaction reward handles imperfect references:} 
Reference motions sometimes fails to precisely establish a grasp due to imperfect retargeting. Interaction reward enables the policy to deviate from an imperfect reference to achieve a proper grasp. Conversely, without it, the agent rigidly follows the flawed reference and fails to interact (\cref{fig:ablation-contact_rew-policy-suitcase} top).

To further validate that the reward's importance lies in handling imperfection, we use a successful policy to collect a perfect reference motion to train. As shown in \cref{fig:ablation-contact_rew-training} and \cref{fig:ablation-contact_rew-policy-suitcase}, training succeeds even without interaction reward, confirming that the interaction reward specifically addresses challenges posed by imperfect reference motions.

\textbf{Interaction reward guides precise contact locations:} 
Pushing box requires the policy to accurately position its L-shaped end effector on the box's edge. 
Without interaction reward (\cref{fig:ablation-contact_rew-policy-push_box} right), the policy fails to achieve such precision. It frequently places end effectors on the vertical surface of the box, which results in highly unstable contact.
This demonstrates interaction reward is essential for learning policies that can maintain stable and precise contact.

\clearpage

\subsection{Residual Action Space}



\begin{figure}[b]
    \centering
    \includegraphics[width=\linewidth]{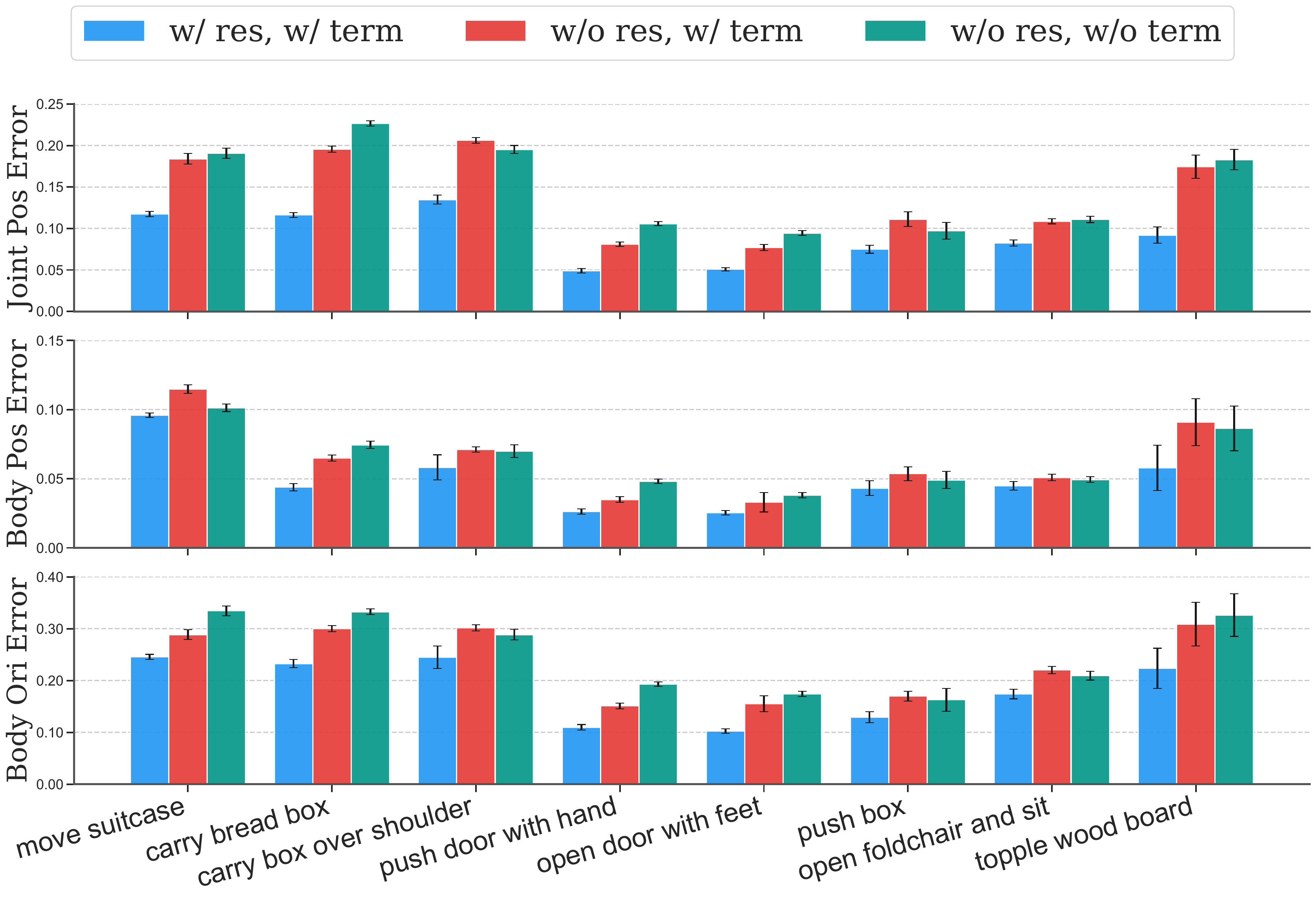}
    \caption{\textbf{Joint and body tracking errors across 8 tasks.} Incorporating a residual action space (\resterm) consistently achieves the lowest tracking errors. Removing it (\noresterm{} and \noresnoterm{}) overall increases the tracking errors. Early termination based on body tracking error (\noresterm{}) marginally decreases tracking error.}

    \label{fig:ablation-exploration-barplot}
\end{figure}

    

We study the impact of two design choices on exploration and convergence: \textbf{residual action space} and \textbf{body tracking error based termination} (\cref{tab:termination_conditions}, \emph{Body Local Pose Error}). We compare three variants for training:
\begin{itemize}
    \item \resterm: Full method that incorporates both residual action space and body tracking error termination.
    \item \noresterm: Removes residual action space; policy outputs are defined relative to the default joint position instead of reference joint position.
    \item \noresnoterm: Further disables body tracking error termination. Root error termination is retained in training to avoid ineffective samples of the robot lying on the ground.
\end{itemize}
Body tracking error based termination is removed for evaluations in this experiment.

\begin{figure}[h!]
    \centering
    \begin{subfigure}{\linewidth}
        \includegraphics[width=\linewidth]{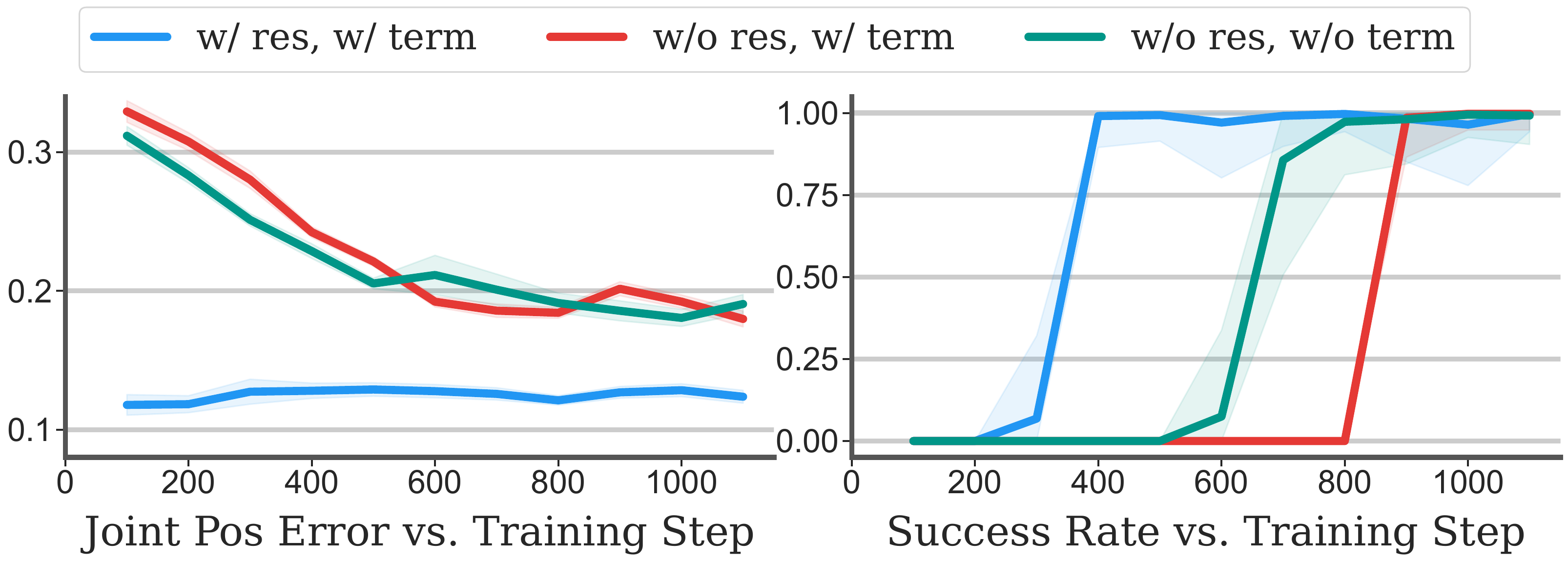}
        \caption{\textbf{Residual actions enable stable and efficient training.} 
        With residual action space, the policy achieves low and stable tracking errors from the beginning, and quickly converges to a high success rate. Policies without it start with higher errors and converge much slower.}
        \label{fig:ablation-exploration-training}
    \end{subfigure}
    \begin{subfigure}{\linewidth}
        \centering
        \includegraphics[width=0.85 \linewidth]{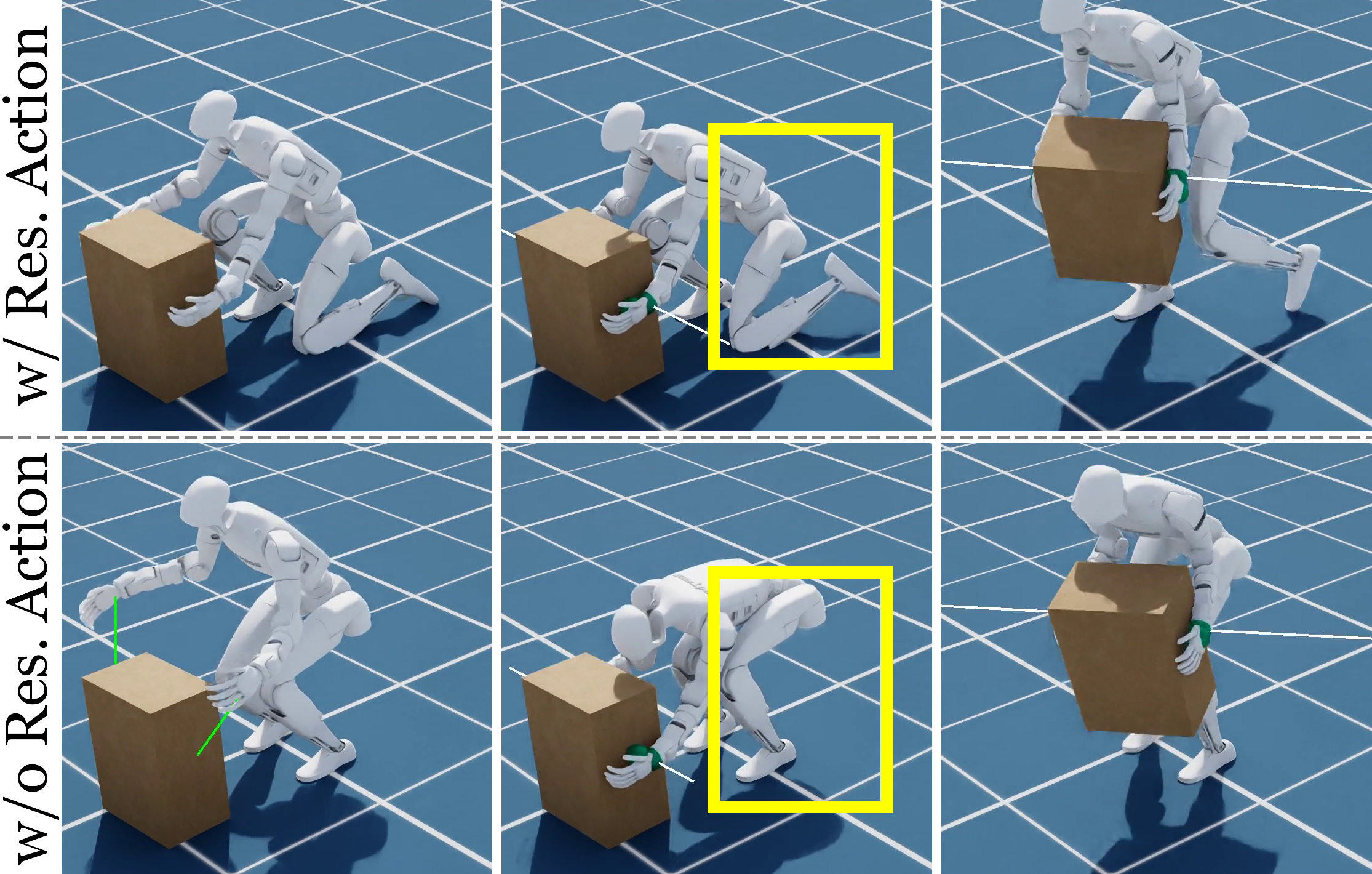}
        \caption{\textbf{Residual actions enable the learning of challenging poses.} With residual actions (top row), the policy successfully executes a kneeling motion to grasp the suitcase. Without them (bottom row), it converges to a suboptimal strategy: keeping both feet flat on the ground and compensates with more waist bending.}
        \label{fig:ablation-exploration-policy}
    \end{subfigure}
    \begin{subfigure}{\linewidth}
        \centering
        \includegraphics[width=0.85 \linewidth]{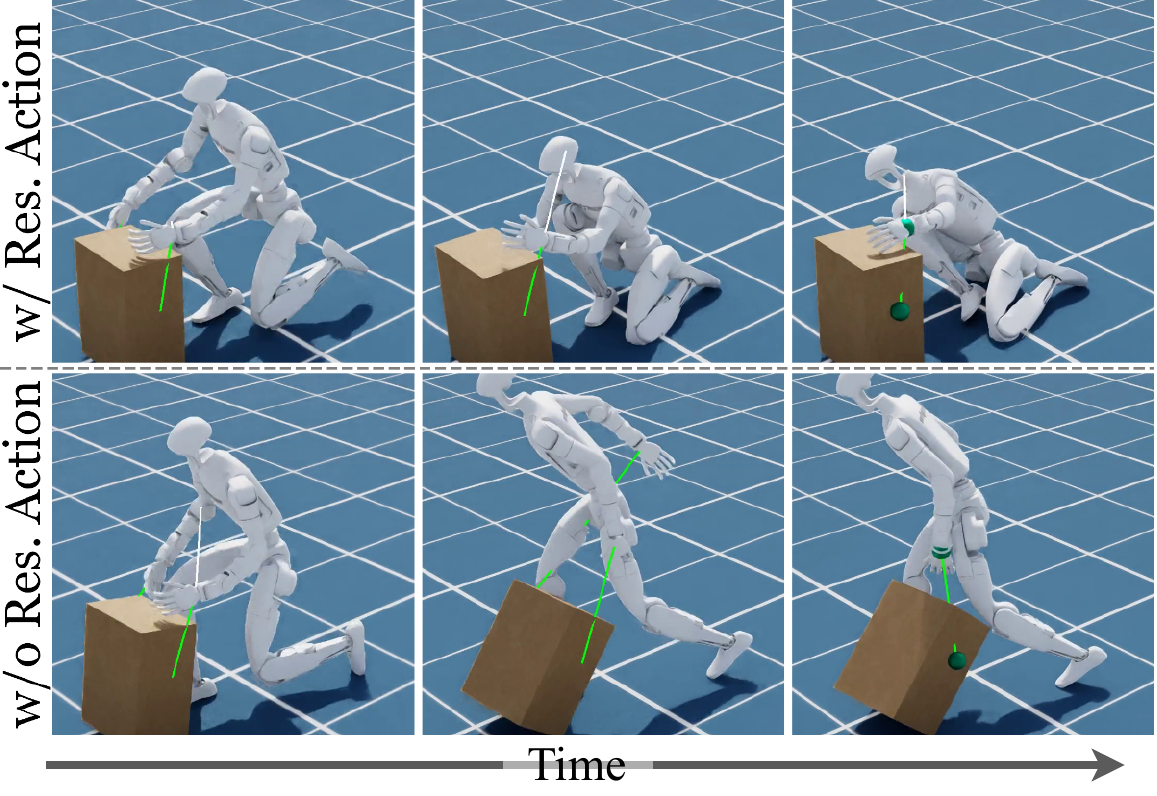}
        \caption{\textbf{Residual actions anchor exploration to the reference motion.}
        When initialized in a kneeling pose, the residual policy (top row) explores around the reference pose. In contrast, the standard policy's exploration is centered on the default pose, causing the robot to abruptly "pop up" and immediately lose balance (bottom row), which generates uninformative training data.}
        \label{fig:ablation-exploration-initial}
    \end{subfigure}

    \caption{Ablation study on residual action space and early termination for body tracking errors.}
    \label{fig:ablation-exploration}
\end{figure}

As shown in \cref{fig:ablation-exploration-barplot}, the full method (\resterm) consistently achieves the lowest joint and body tracking errors across 8 tasks. Residual action space allows faster convergence and higher performance: without it (\noresterm{} and \noresnoterm{}), training converges much slower and fails to reach the same level of performance (\cref{fig:ablation-exploration-training}).  

For move suitcase task, without both components (\noresnoterm{}) the policy cannot learn the intended kneeling motion \cref{fig:ablation-exploration-policy}. Instead, it converges to a suboptimal strategy of bending at the waist while keeping both feet flat.  

This gap arises because residual action space grounds exploration around the reference motion. Without it, actions are defined relative to the default standing pose, thus initial exploration is centered around the standing pose; 
when an episode is initialized from kneeling, this leads to abrupt ``pop up'' behaviors (\cref{fig:ablation-exploration-initial}, bottom row), producing low-quality samples. 
In contrast, residual policies explore locally around the reference pose (top row), enabling stable training, faster convergence, and successful acquisition of challenging skills.



\section{Limitations and Future Directions}
\label{sec:limitations}

Our framework, \method{}, enables humanoids to acquire diverse object interaction skills from human videos. While we have demonstrated its effectiveness across 14 simulated tasks, two key limitations remain:

\textbf{Dependence on Mocap.} The current system relies on ground-truth motion capture data (e.g., object poses). A critical next step is to develop policies that operate directly from on-board sensing modalities, such as cameras, to enable deployment in uninstrumented environments.

\textbf{One Policy per Skill.} At present, a separate specialist policy is trained for each task. An important future direction is to leverage the data from multiple skills to train a unified generalist model, capable of performing a wide range of interactions.

\section{Acknowledgement}

We would like to thank Guanqi He for constructing the door setup. We are also grateful to Haotian Lin, Chaoyi Pan, Yuanhang Zhang, and Wenli Xiao for their valuable discussions. Guanya Shi holds concurrent appointments as an Assistant Professor at Carnegie Mellon University and as an Amazon Scholar. This paper describes work performed at Carnegie Mellon University and is not associated with Amazon.

\printbibliography








\end{document}